\documentclass[final]{cvpr}

\usepackage{times}
\usepackage{epsfig}
\usepackage{graphicx}
\usepackage{amsmath}
\usepackage{amssymb}
\usepackage{float}
\usepackage{colortbl}
\usepackage{booktabs}
\usepackage{multirow}

\usepackage[pagebackref=true,breaklinks=true,colorlinks,bookmarks=false]{hyperref}

\def\cvprPaperID{****} 
\def\confYear{CVPR 2021}
\begin{document}

\title{\vspace{-10mm}Self-training Guided Adversarial Domain Adaptation For Thermal Imagery}

\author{Ibrahim Batuhan Akkaya\textsuperscript{1,2}\thanks{indicates equal contribution} \qquad\qquad Fazil Altinel\textsuperscript{1}\footnotemark[1] \qquad\qquad Ugur Halici\textsuperscript{2,3}\\
\textsuperscript{1}Research Center, Aselsan Inc., Turkey\\
\textsuperscript{2}Department of Electrical and Electronics Engineering, Middle East Technical University, Turkey\\
\textsuperscript{3}NOROM Neuroscience and Neurotechnology Excellency Center, Turkey\\
{\tt\small \{ibakkaya, faltinel\}@aselsan.com.tr}, {\tt\small halici@metu.edu.tr}
}

\maketitle

\begin{abstract}
Deep models trained on large-scale RGB image datasets have shown tremendous success. It is important to apply such deep models to real-world problems. However, these models suffer from a performance bottleneck under illumination changes. Thermal IR cameras are more robust against such changes, and thus can be very useful for the real-world problems. In order to investigate efficacy of combining feature-rich visible spectrum and thermal image modalities, we propose an unsupervised domain adaptation method which does not require RGB-to-thermal image pairs. We employ large-scale RGB dataset MS-COCO as source domain and thermal dataset FLIR ADAS as target domain to demonstrate results of our method. Although adversarial domain adaptation methods aim to align the distributions of source and target domains, simply aligning the distributions cannot guarantee perfect generalization to the target domain. To this end, we propose a self-training guided adversarial domain adaptation method to promote generalization capabilities of adversarial domain adaptation methods. To perform self-training, pseudo labels are assigned to the samples on the target thermal domain to learn more generalized representations for the target domain. Extensive experimental analyses show that our proposed method achieves better results than the state-of-the-art adversarial domain adaptation methods. The code and models are publicly available.\footnote[1]{\href{https://github.com/avaapm/SGADA}{https://github.com/avaapm/SGADA}}
\end{abstract}

\section{Introduction}

\begin{figure}[t]
\centering
    \scalebox{0.85}{\includegraphics{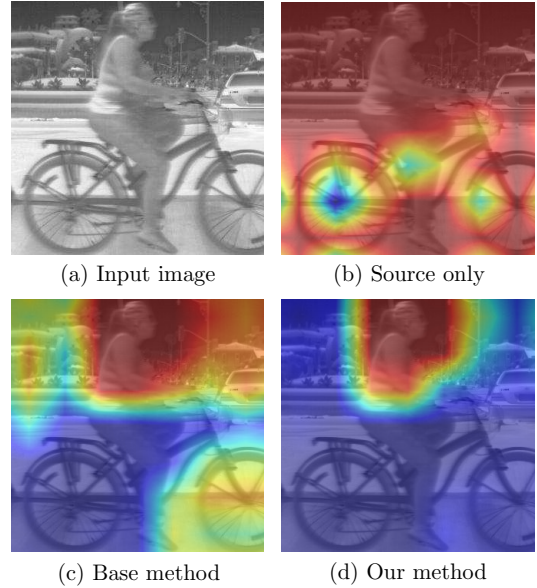}}
    \vspace{3mm}
    \caption{Visualization of class activation maps on a target domain image using occlusion sensitivity \cite{zeiler2014visualizing}. Given a person image (a), our proposed model (d) activates semantically more meaningful parts of the image compared to our base method (c) \cite{adda}, while the model trained on only source domain images (b) misclassifies the image as bicycle and activates wrong regions.  \textit{Best viewed in color}.}
    \label{fig:activation_maps}
\end{figure}

Recently, significant improvements have been made by using RGB images for image classification and detection problems \cite{rcnn, resnet, alexnet, yolo, fasterrcnn}. The state-of-the-art methods have been trained on large-scale RGB datasets such as MS-COCO \cite{mscoco}, ImageNet \cite{imagenet_cvpr09}, Pascal-VOC \cite{pascalvoc}, etc. However, low-lighting conditions hinder current state-of-the-art deep learning methods trained on visible spectrum images from performing well on computer vision tasks such as image classification, object detection, etc. Since thermal IR cameras are more robust against these conditions, exploiting them is useful for real-world applications. Therefore, usage of thermal IR cameras has become more common in the tasks related to autonomous driving, military operations, security surveillance, etc. Since such large-scale thermal datasets are not publicly available, it still remains an important challenge to achieve same level of performance on thermal image datasets. Therefore, exploiting complementary information offered by visible spectrum images is a straightforward technique to improve performance of the methods which work on thermal images for classification and detection problems. Unfortunately, recent studies demonstrated that performance of a deep model well-trained on visible spectrum images may significantly drop when applying to thermal images \cite{devaguptapu2019borrow, guan2019unsupervised, guo2019domain, kaist, BMVC2016_73}. 

Since deep networks are sensitive to domain shift, a deep model trained on a large amount of labeled source domain data may fail at generalizing to unlabeled target domain data which are not similar to source domain data. To overcome these issues, unsupervised domain adaptation (UDA) aims to learn a model which maps both domains into a common feature space without requiring image pairs. Among the recent UDA methods, adversarial domain adaptation methods have become popular \cite{dann, tat, cdan, adda}. These approaches incorporate adversarial learning as a two-player game similar to generative adversarial networks (GANs) \cite{gan}. Adversarial domain adaptation methods utilize a domain discriminator to distinguish source domain from target domain and a feature extractor to learn domain invariant representations to fool the domain discriminator. By learning domain invariant feature representations, adversarial domain adaptation methods assume that a classifier trained on source domain features is able to successfully classify target domain samples as well. 

In this paper, we propose an unsupervised adversarial domain adaptation method to align source and target domain distributions as described in Section \ref{sec:method}. We employ Adversarial Discriminative Domain Adaptation (ADDA) \cite{adda} method as our base method. Although ADDA and other adversarial domain adaptation methods have achieved successful results, these methods face a major generalization limitation. The limitation is that even though the distributions are aligned by learning domain invariant representations with a feature extractor, theoretically, the classifier may not work well on the target domain as shown in \cite{ben2010theory}. Therefore, learning discriminative representations for the unlabeled target domain is a difficult problem.

\begin{figure*}[ht]
\centering
    \scalebox{0.77}{\includegraphics{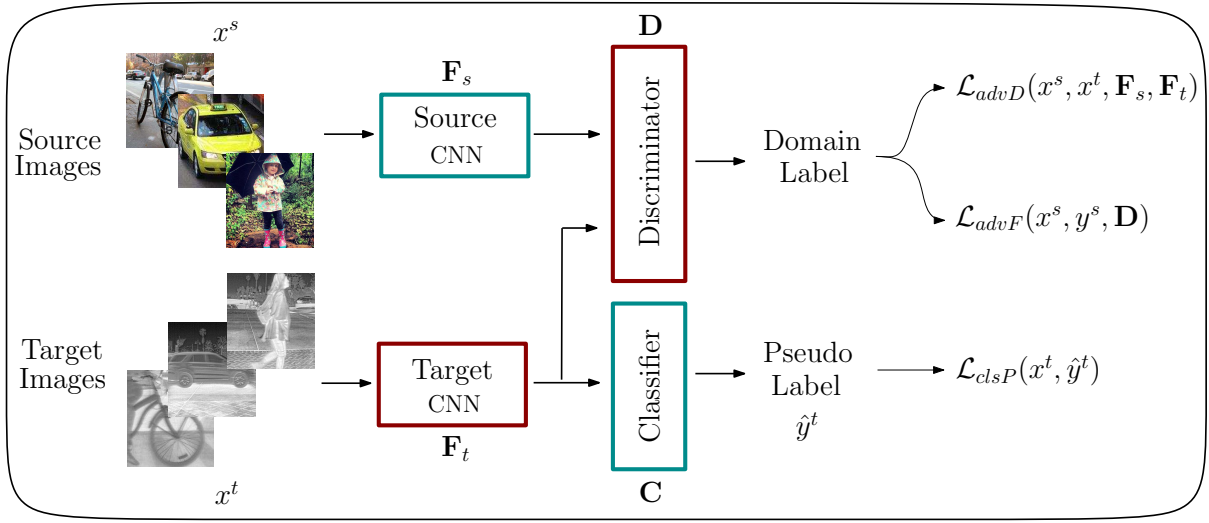}}
    \vspace{3mm}
    \caption{An overview of our proposed self-training guided adversarial domain adaptation (SGADA) method. Pseudo-labels generated after our methods' warm-up phase are assigned to target thermal images. Then, the target CNN $ \mathbf{F}_t $ is trained using the pseudo-labels. The classifier $ \mathbf{C} $ and the source CNN $ \mathbf{F}_s $ are reused from our base model, and thus they are fixed. Target feature representations are learned by updating parameters of the target CNN $ \mathbf{F}_t $ with respect to losses generated by the discriminator $ \mathbf{D} $ and the classifier $ \mathbf{C} $. Blue boxes indicate fixed network parameters while red boxes indicate trainable network parameters. \textit{Best viewed in color}.}
    \label{fig:overview}
\end{figure*}

Based on the assumption of self-training, a classifiers' own high-confidence predictions are correct \cite{zhu2005semi}. Since we assume that the predictions are mostly correct, exploiting the samples with high confidence values and retraining the classifier further improves the performance of the classifier. To this end, recent adversarial domain adaptation methods proposed to use pseudo-labels obtained from a classifier and retrain the model using the pseudo-labeled samples \cite{saito2017asymmetric, xie2018learning, zhang2018collaborative}. With these in mind, in this study, we propose a self-training guided adversarial domain adaptation (SGADA) method to overcome the generalization problems of adversarial domain adaptation methods (Figure \ref{fig:overview}). To perform self-training, pseudo labels obtained after warm-up phase of our method are assigned to the samples on the target domain to learn more generalized representations for the target domain. Pseudo-labels are assigned if confidences of the classifier trained on source domain and the domain discriminator reaches to threshold values for a target domain sample.

Our proposed method makes use of features obtained from visual spectrum images to improve classification performance on thermal domain. Moreover, our method does not need paired samples of RGB and thermal datasets. In order to train and test our proposed method, we use large-scale RGB dataset MS-COCO \cite{mscoco} and thermal imagery dataset FLIR ADAS \cite{flir}. We evaluate the proposed method quantitatively and qualitatively. We demonstrate our methods' success compared to the state-of-the-art unsupervised domain adaptation methods in Section \ref{sec:experiments}. The results show that our method improves the performance of our base model and outperforms the state-of-the-art methods. Moreover, Figure \ref{fig:activation_maps} depicts that given a thermal image, our method classifies the image correctly by activating semantically more meaningful regions compared to our base method ADDA \cite{adda} and the model trained only on source domain data.
\vspace{2mm}

Effective classification for imbalanced data is an important field of research since class imbalance exists in many real-world applications \cite{buda2018systematic, classimbalance}. Therefore, it is important to address the class imbalance problem. In our experimental studies (Section \ref{sec:experiments}), we show that class imbalanced datasets cause UDA methods to over-classify the majority category. We show that our proposed method achieves better results compared to the state-of-the-art UDA methods when class imbalance exists.
\vspace{2mm}

Our contributions are summarized as follows:
\begin{itemize}
\item We demonstrate the efficacy of combining visible spectrum and thermal image modalities by using unsupervised domain adaptation without requiring RGB-to-thermal image pairs.
\item We propose a self-training guided adversarial domain adaptation method for thermal imagery. In order to learn more generalized feature representations for target thermal domain, we employ pseudo-labels generated by the classifier trained on RGB images and the discriminator, and train our model with these pseudo-labels.
\item In order to demonstrate results of our method, we employ the large-scale RGB dataset MS-COCO as source domain and the thermal dataset FLIR ADAS as target domain. Extensive experimental analyses show that our proposed method outperforms the state-of-the-art unsupervised domain adaptation methods.
\end{itemize}

\section{Related Work}

By using RGB images, deep neural networks have gained popularity on computer vision tasks such as object detection, classification etc. Although significant improvements have been accomplished by using visible spectrum images, it still remains a critical problem to train a deep model robust to real-world problems, e.g. low-lighting conditions. To overcome these problems, thermal imagery has been used for object detection and classification problems \cite{guo2019domain, kaist, saponaro2015material}.

Some of recent studies investigated the effects of combining RGB and thermal images to the performance on object detection problem \cite{devaguptapu2019borrow, guo2019domain, kaist, BMVC2016_73}. Since large-scale thermal datasets are not publicly available, in this study, we exploit complementary information offered by visible spectrum images to improve classification performance on thermal imagery without requiring RGB-to-thermal image pairs. We propose an unsupervised domain adaptation method in order to investigate efficacy of combining visible spectrum and thermal image modalities. 

Numerous recent methods attempted to address domain adaptation problem. Recently, Generative Adversarial Networks (GANs) \cite{gan} inspired the field of domain adaptation, and thus deep adversarial domain adaptation methods have become popular \cite{pixelda, dann, tat, cdan, adda}.

Feature-level adversarial domain adaptation methods incorporate a domain discriminator to distinguish source and target domains while feature extractor learns features to fool the discriminator. Ganin et al. \cite{dann} proposed a gradient reversal layer to learn a feature extractor which generates features that maximize domain discriminator loss while minimizing label prediction loss. More recently, Tzeng et al. \cite{adda} proposed a method to learn a discriminative mapping of target images to source feature space by fooling a domain discriminator which distinguishes the encoded target images from source samples. Many recent works employ adversarial training paradigm in their domain adaptation procedure \cite{tat, cdan}. Although feature-level adversarial domain adaptation methods have accomplished successful empirical results, these methods suffer from a major limitation. As shown in \cite{ben2010theory}, even if a feature extractor is well learned to generate domain invariant features, theoretically, the classifier may not work well on the target domain. Therefore, learning discriminative representations for the unlabeled target domain is considered difficult. 

On the other hand, pixel-level adversarial domain adaptation methods translate source domain data into target domain data or vice versa by using image-to-image translation \cite{itoi}. Bousmalis et al. \cite{pixelda} proposed an approach to learn a transformation in pixel-level from one domain to the other. Inspired by CycleGAN \cite{cyclegan}, Hoffman et al. \cite{cycada} proposed CyCADA to increase semantic consistency of the image translation to improve the pixel-level methods. Even though pixel-level adversarial domain adaptation studies present remarkable results, image-to-image translation sometimes performs poorly on the datasets which have objects with many complex structures.

To overcome the limitations of adversarial domain adaptation methods, recent studies propose to directly deal with relationship between decision boundary and learned feature representations \cite{dta, mcdda}. Saito et al. \cite{mcdda} introduced to use a minimax training method to push target feature distributions away. Lee et al. \cite{dta} proposed to exploit adversarial dropout mechanism to learn more discriminative features by enforcing cluster assumption \cite{chapelle2005semi}. However, our experimental studies show that these methods and aforementioned adversarial domain adaptation methods have a drawback: Class imbalanced datasets \cite{buda2018systematic, classimbalance} lead to a performance drop for these methods.

We employ a self-training guided adversarial domain adaptation method to deal with the generalization problems of adversarial domain adaptation methods for thermal imagery. To the best of our knowledge, there is no self-training guided domain adaptation study in the literature of thermal image classification. Self-training is a technique to assign pseudo-labels to unlabeled samples using predictions of a classifier and retrain the model including the pseudo-labeled samples. Based on the assumption of self-training, a classifiers' own high-confidence predictions are correct \cite{zhu2005semi}. Recent adversarial domain adaptation methods proposed to use pseudo-labels \cite{saito2017asymmetric, xie2018learning, zhang2018collaborative}. In our experiments, we show that our self-training guided method performs better than previous domain adaptation methods under class imbalance problem by learning more generalized representations for target thermal domain.

\section{Proposed Method}
\label{sec:method}

\begin{figure}[t]
\centering
    \scalebox{0.70}{\includegraphics{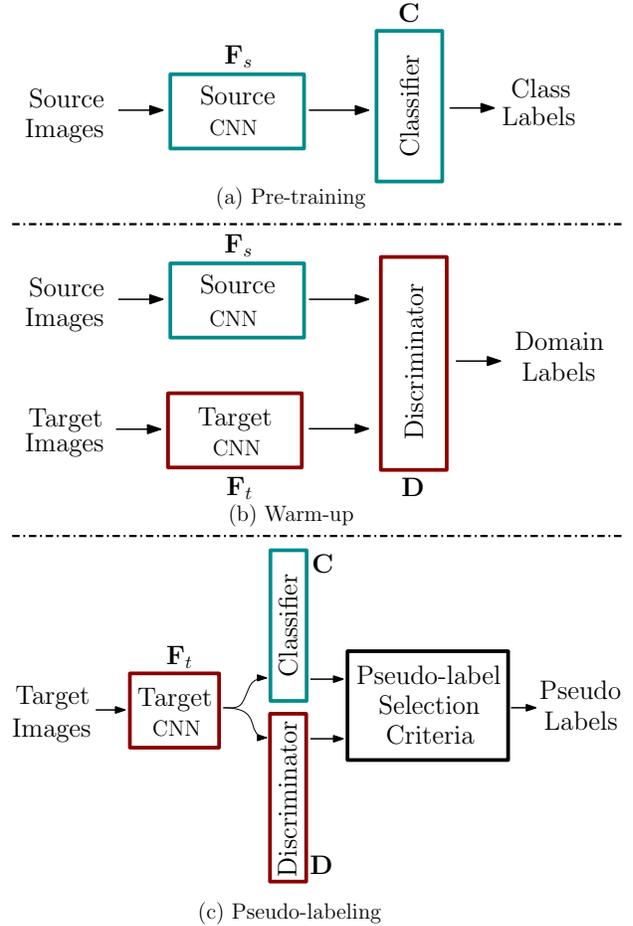}}
    \vspace{2mm}
    \caption{Illustration of our pseudo-label generation mechanism.}
    \label{fig:adda}
\end{figure}

Our proposed self-training guided adversarial domain adaptation method is illustrated in Figure \ref{fig:overview}. Before performing self-training, we extract pseudo-labels for the target domain samples. The pseudo-label extraction mechanism is depicted in Figure \ref{fig:adda}. 

First, a feature extractor $ \mathbf{F}_s $ and a classifier $ \mathbf{C} $ are trained on source domain using labeled source domain RGB images (Figure \ref{fig:adda}-(a)). This step is named as pre-training. After this step, the classifier network $ \mathbf{C} $ can successfully classify the source domain images by exploiting the features which are extracted by the source Convolutional Neural Network (CNN) $ \mathbf{F}_s $. After the training on the source domain, we perform the second step: the warm-up phase for pseudo-label generation (Figure \ref{fig:adda}-(b)). In this step, we fix the parameters of the feature extractor $ \mathbf{F}_s $ trained on the source domain. A target specific feature extractor $ \mathbf{F}_t $ is learned in an unsupervised manner. By performing this step, features extracted from the source domain and the target domain are aligned with adversarial training. Therefore, we can use the classifier $ \mathbf{C} $ trained on the source domain to classify target domain samples. We perform aforementioned two steps by following the training process of our base method ADDA \cite{adda}. In the last step, we fix the parameters of the feature extractor $ \mathbf{F}_t $ trained on the target domain, the classifier $ \mathbf{C} $ trained on the source domain and the discriminator $ \mathbf{D} $. Then, we obtain predictions from the classifier and confidences from both the classifier and the discriminator for the target domain samples.

Once the predictions and the confidences are obtained, we utilize the predictions to give pseudo-labels for target domain samples using the confidences obtained from the classifier $ \mathbf{C} $ and the discriminator $ \mathbf{D} $ as shown in Figure \ref{fig:adda}-(c). We use prediction of the classifier for a target sample if classifiers' confidence value is higher than a threshold and domain label prediction of discriminator is close to source domain. That is, we can use the prediction of the classifier if the discriminator incorrectly classifies target samples. By using this pseudo-label selection mechanism, intuitively, we select samples with feature representations which are close to data with known labels. Next, as illustrated in Figure \ref{fig:overview}, we train our proposed method using extracted pseudo-labels.

A general definition of unsupervised domain adaptation, and self-training guided adversarial domain adaptation procedures of our proposed method are described in Section \ref{sec:uda} and \ref{sec:selftraining}, respectively.

\subsection{Unsupervised Domain Adaptation}
\label{sec:uda}

In the general definition of unsupervised domain adaptation (UDA) problem, we are given $ n_s $ labeled samples from a source domain $ \mathcal{D}_s = \{(x^s_i, y^s_i)\}_{i=1}^{n_s} $ and $ n_t $ unlabeled samples from a target domain $ \mathcal{D}_t = \{(x^t_j)\}_{j=1}^{n_t} $. The goal of UDA problem is to learn a feature extractor $ \mathbf{F}_t $ for the target domain and a classifier $ \mathbf{C}_t $ which correctly classifies the features. It is not possible to perform supervised training since there is no labeled samples in the target domain. Therefore, UDA learns to adapt the source feature extractor $ \mathbf{F}_s $ and the source classifier $ \mathbf{C}_s $ to be able to use them on target domain.

\subsection{Self-training Guided Adversarial Domain Adaptation (SGADA)}
\label{sec:selftraining}

The task of adversarial domain adaptation methods is to adversarially align source and target domain representations. For this purpose, adversarial domain adaptation methods propose to reduce the gap between $ \mathbf{F}_s(x^s) $ and $ \mathbf{F}_t(x^t) $. Thus, the classifier $ \mathbf{C}_s $ trained on source domain can be applied to the representations on target domain, and necessity to train a separate $ \mathbf{C}_t $ can be eliminated. As a result, we obtain $ \mathbf{C}=\mathbf{C}_s=\mathbf{C}_t $ \cite{adda}. We employ the feature extractor $ \mathbf{F}_s $ and the classifier $ \mathbf{C} $ which are learned during the warm-up phase. In this subsection, we elaborate our training scheme shown in Figure \ref{fig:overview}.

We use the following loss function for the domain discriminator $ \mathbf{D} $ which distinguishes source domain from target domain:

\begin{equation}
\label{eq:discloss}
    \begin{aligned}
        \mathcal{L}_{advD}(x^s, x^t, \mathbf{F}_s, \mathbf{F}_t) = -\frac{1}{n_s}\sum_{i=1}^{n_s}\log[\mathbf{D}(\mathbf{F}_s(x_i^s))]\\ -\frac{1}{n_t}\sum_{i=1}^{n_t}\log[1-\mathbf{D}(\mathbf{F}_t(x_i^t))].
    \end{aligned}
\end{equation}

Given source images $ x^s $ and target images $ x^t $, we update the parameters of the domain discriminator $ \mathbf{D} $ with respect to outputs of the feature extractors $ \mathbf{F}_s $ and $ \mathbf{F}_t $. While updating the parameters, we fix and reuse the source feature extractor $ \mathbf{F}_s $ which is trained in the pre-training step of our pseudo-label generation.
\vspace{2mm}

We employ two loss functions to train the target CNN $ \mathbf{F}_t $: adversarial loss $ \mathcal{L}_{advF} $ and self-training loss $ \mathcal{L}_{clsP} $. The adversarial loss is formulated as follows: 

\begin{equation}
\label{eq:ganobj}
    \mathcal{L}_{advF}(x^s, y^s, \mathbf{D}) = \frac{1}{n_t}\sum_{i=1}^{n_t}\log[\mathbf{D}(\mathbf{F}_t(x_i^t))].
\end{equation}

Note that we reuse the parameters of the source feature extractor $ \mathbf{F}_s $ from the previous step to initialize $ \mathbf{F}_t $.
\vspace{2mm}

We exploit pseudo-labeled target domain samples to perform self-training guided adversarial learning. After the learning of the classifier $ \mathbf{C} $ and the domain discriminator $ \mathbf{D} $ is completed during the warm-up phase, we obtain predictions from the classifier and confidences for these predictions. Given an unlabeled target domain sample, if confidence of the classifier $ \mathbf{C} $ is higher than pre-defined threshold and the domain discriminator $ \mathbf{D} $ classifies the sample as source domain, we include the sample during our self-training guided adversarial domain adaptation step. Also, if the domain discriminator $ \mathbf{D} $ classifies the sample as target domain with a confidence lower than a pre-defined threshold, we assign the pseudo-label $ \hat{y}^t $ generated by the classifier $ \mathbf{C} $ to the sample as well. By using $ \hat{n}_t $ pseudo-labeled samples on target domain, we aim to train a target specific feature extractor (Figure \ref{fig:overview}). We use the new self-training loss function for our proposed method:

\begin{equation}
    \mathcal{L}_{clsP}(x^t, \hat{y}^t) = \frac{1}{\hat{n}^t}\sum_{i=1}^{\hat{n}^t}\ell_{ce}(\mathbf{C}(\mathbf{F}_t(x_i^t)), \hat{y}_i^t).
\end{equation}

The overall objective function to train our proposed method SGADA is defined as:

\begin{equation}
\label{eq:allobj}
\begin{aligned}
    \min_{\mathbf{D}}\mathcal{L}_{advD}(x^s, x^t, \mathbf{F}_s, \mathbf{F}_t) 
    \\\min_{\mathbf{F}_t}\mathcal{L}_{advF}(x^s, y^s, \mathbf{D})
    \\+\lambda\mathcal{L}_{clsP}(x^t, \hat{y}^t),
\end{aligned}
\end{equation}
where $ \lambda $ is a trade-off parameter. We set the trade-off parameter $ \lambda $ and thresholds based on validation split (see Section \ref{sec:implementation} for further details).

\section{Experiments}
\label{sec:experiments}

We perform extensive evaluations and compare our proposed method with several state-of-the-art unsupervised domain adaptation methods.

\subsection{Datasets}

We prepare a new RGB-to-thermal domain adaptation setting for classification using FLIR ADAS \cite{flir} as thermal dataset and MS-COCO \cite{mscoco} as visible spectrum dataset for our experimental studies.
\vspace{2mm}

\begin{figure*}[ht]
\centering
    \scalebox{1.06}{\includegraphics{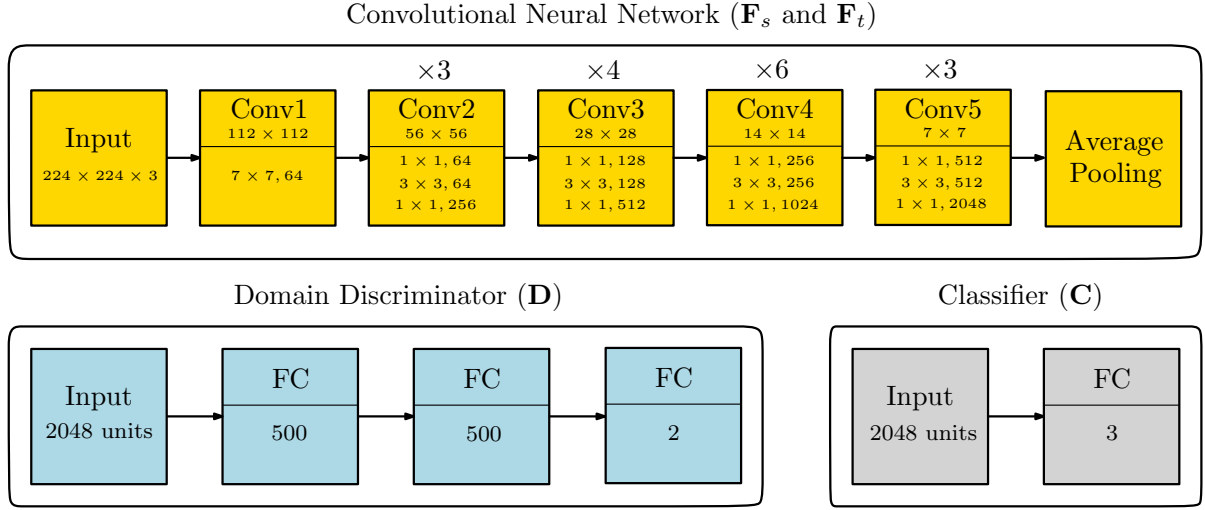}}
    \vspace{3mm}
    \caption{The network architectures used for our experimental analyses.}
    \label{fig:architecture}
\end{figure*}

\begin{figure}[t]
\centering
\scalebox{0.66}{\includegraphics{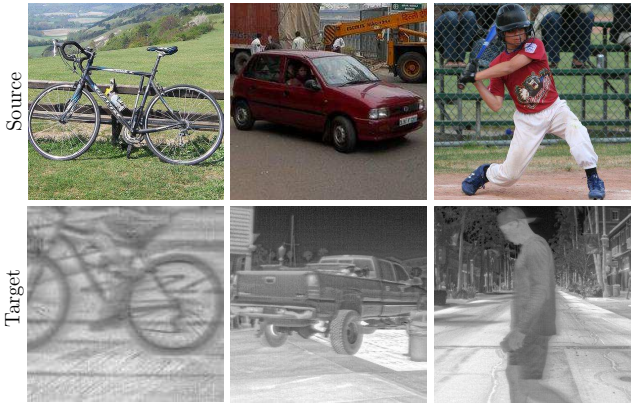}}
\vspace{2mm}
\caption{Example images from MS-COCO dataset \cite{mscoco} and FLIR ADAS thermal dataset \cite{flir}. \textit{Best viewed in color}.}
\label{fig:dataset}
\end{figure}

FLIR ADAS \cite{flir} consists of 9,214 thermal images with bounding box annotations. Each image has a resolution of 640 $ \times $ 512 and obtained from FLIR Tau2 camera. 60\% of the images are captured during daytime and the remaining 40\% of the images are captured during night. The dataset provides both visible spectrum (RGB) images and thermal images. We consider only the thermal images of the dataset for our experiments. We use the training and test splits as suggested in the dataset documentation for our experiments. The objects in the dataset are classified into four categories i.e. bicycle, car, dog, and person. However, the dog class has very few annotations. Therefore, the dog class is not considered in our experimental studies. We crop square images using bounding box annotations for objects. After objects are extracted, we resize the images to 224 $ \times $ 224. Finally, our thermal dataset consists of 4,137 samples of bicycle, 43,734 samples of car, and 26,294 samples of person images. Example images from FLIR ADAS dataset are shown in the second row of Figure \ref{fig:dataset}.

\begin{figure*}[ht]
\centering
    \scalebox{0.99}{\includegraphics{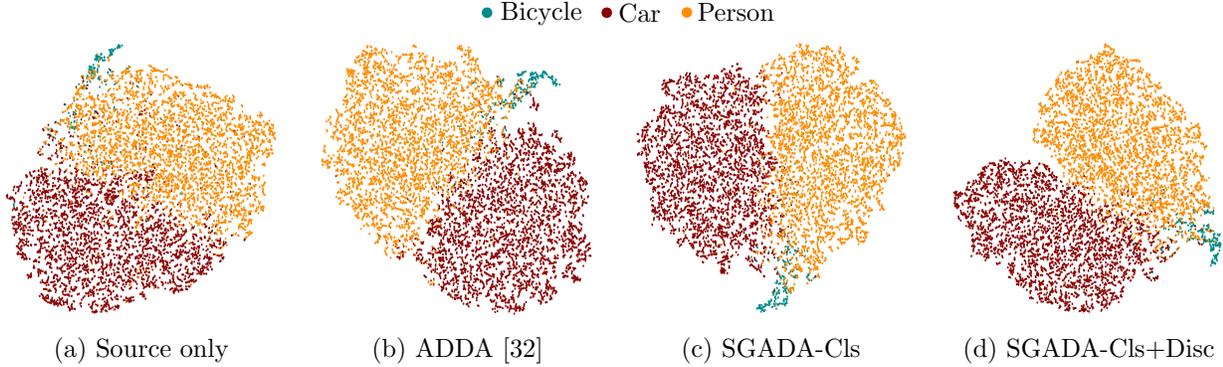}}
    \vspace{3mm}
    \caption{The t-SNE visualization of network activations on target thermal domain generated by source only model (a), our base method ADDA \cite{adda} (b), our proposed method SGADA with classifier confidences only (c), and our proposed method SGADA with classifier and discriminator confidences (d). \textit{Best viewed in color}.}
    \label{fig:tsne}
\end{figure*}

Our proposed method incorporates publicly available large-scale visible spectrum datasets to improve classification performance on thermal dataset. Therefore, we consider using an RGB dataset which includes the same classes as FLIR dataset \cite{flir} (bicycle, car, and person). For this purpose, we use MS-COCO dataset \cite{mscoco} as our visible spectrum dataset. In the first row of Figure \ref{fig:dataset}, we show some example images from MS-COCO dataset. MS-COCO dataset contains 91 object categories (airplane, bicycle, bird, car, person, etc.). In total, there are 123,287 images and around 886,000 bounding boxes. 118,287 of the images are for training while 5,000 of the images are for validation split. We apply standard training and test splits as provided in the dataset documentation for our experiments. We use only bicycle, car, and person classes to match with our thermal dataset. We cropped the annotated objects with the same procedure as applied to FLIR dataset. Once objects are extracted, we resize the images to 224 $ \times $ 224. Our visible spectrum image dataset extracted from MS-COCO consists of 5,732 samples of bicycle, 38,453 samples of car, and 209,162 samples of person images.

\subsection{Implementation Details}
\label{sec:implementation}

For our experiments, we used same training procedure with ADDA \cite{adda}. For a fair comparison with other methods, we employed ResNet-50 \cite{resnet} pre-trained on ImageNet \cite{imagenet_cvpr09} as backbone for all methods. Our network architectures are given in Figure \ref{fig:architecture}. The architecture of our feature extractors (source CNN $ \mathbf{F}_s $ and target CNN $ \mathbf{F}_t $) is the ResNet-50 without the last fully connected (FC) layer. In the figure, each convolutional residual unit is depicted with the size of filters at the top and the outputs of each convolutional layer at the bottom. The notation k $ \times $ k, n in the convolutional layer blocks represents a filter of size k and n channel. The number on the top of the convolutional layer blocks denotes the number of repetition for the unit. The domain discriminator $ \mathbf{D} $ consists of three FC layers: two consecutive hidden units of 500 neurons and the discriminator output. The classifier $ \mathbf{C} $ has only one FC layer.
\vspace{2mm}

To train our method, we set batch size to 32. Number of epochs was set to 15 in our experiments. Parameters were updated using ADAM optimization algorithm \cite{adam}. For pre-training of our method on source domain only, we set learning rate to 5e-4. For adversarial adaptation step of our method, we set learning rate as 1e-5 and discriminator learning rate as 1e-3. We used same learning rates as used in the previous step for self-training guided adversarial adaptation step of our method. We set $ \lambda $ value as 0.7 and threshold value as 0.87 for our method with classifier confidences only (SGADA-Cls). For our method with classifier and discriminator confidences (SGADA-Cls+Disc), we used same learning rates as used in the previous step, and $ \lambda $ value of 0.25, classifier threshold value of 0.79, and discriminator threshold value of 0.87. We used the same experimental settings for training and testing. We exploit classification accuracy to compare our proposed method with other methods.
\vspace{2mm}

We implemented our proposed method using PyTorch framework \cite{pytorch}. Implementation details, models, and the code were made publicly available at \href{https://github.com/avaapm/SGADA}{https://github.com/avaapm/SGADA}.

\subsection{Evaluation of SGADA}

In our experiments, we select visual spectrum (RGB) domain as the source domain, and thermal domain as the target domain. As a general practice in the field of domain adaptation, we denote \textit{source only} as the target domain performance of a model trained using only source domain images, and \textit{target only} as that of a model trained on the target domain. Performances of \textit{source only} and \textit{target only} models serve as baselines for the lower and upper bound performances.

\begin{table}[ht]
\renewcommand{\arraystretch}{1.3}
\caption{Per-class classification performance comparison.}
\label{table:results}
\vspace{3mm}
\centering
\scalebox{1.07}{
\begin{tabular}{@{}l c c c >{\columncolor[gray]{0.8}}c}
\toprule
    Method & \rotatebox{90}{Bicycle } & \rotatebox{90}{Car } & \rotatebox{90}{Person } & \rotatebox{90}{\textbf{Average }}\\
\hline
Source only & 69.89 & 83.89 & 86.52 & 80.10\\
Pixel-DA \cite{pixelda} & 62.53 & 89.99 & 76.73 & 76.42\\
DTA \cite{dta} & 75.45 & \textbf{97.65} & 92.45 & 88.52\\
MCD-DA \cite{mcdda} & 81.71 & 94.90 & 91.83 & 89.48\\
DANN \cite{dann} & 78.16 & 95.07 & \textbf{96.24} & 89.82\\
CDAN \cite{cdan} & 78.16 & 97.10 & 94.82 & 90.03\\
ADDA \cite{adda} & 86.67 & 96.95 & 89.10 & 90.90\\
SGADA (ours) & \textbf{87.13} & 94.44 & 92.03 & \textbf{91.20}\\
\hline
Target only & 87.59 & 98.78 & 96.35 & 94.24\\
\bottomrule
\end{tabular}}
\end{table}

\paragraph{Quantitative Analysis.} We compare our proposed method SGADA with several state-of-the-art unsupervised domain adaptation methods: Unsupervised Pixel-Level Domain Adaptation with Generative Adversarial Networks (Pixel-DA) \cite{pixelda}, Drop to Adapt (DTA) \cite{dta}, Maximum Classifier Discrepancy for Unsupervised Domain Adaptation (MCD-DA) \cite{mcdda}, Domain Adversarial Neural Network (DANN) \cite{dann}, Conditional Domain Adversarial Adaptation (CDAN) \cite{cdan}, and Adversarial Discriminative Domain Adaptation (ADDA) \cite{adda}. Since these methods do not consider domain adaptation problem for thermal datasets, there exist no reported results on their paper for our dataset. Therefore, we trained and evaluated all these methods for our dataset. 
\vspace{2mm}

\begin{table*}[ht]
\renewcommand{\arraystretch}{1.3}
\caption{Ablation studies of different pseudo-label selection scenarios for self-training guided adversarial domain adaptation.}
\label{table:confidence}
\vspace{2mm}
\centering
\scalebox{1.03}{
\begin{tabular}{@{}llccc@{}}
\toprule
\multicolumn{2}{l}{}                                             & Bicycle & Car & Person\\
\hline
& Number of samples                        & 3702      & 38657    & 21081\\
\hline
\multirow{3}{*}{Classifier confidences only}          & Number of selected samples       & 3995      & 35494    & 20323  \\
& Number of correctly selected samples & 2901      & 34905    & 18911 \\
& Accuracy of selected samples (\%)   & 72.62     & 98.34   & 93.05 \\
\hline
\multirow{3}{*}{Discriminator confidences only} & Number of selected samples       & 3598      & 36024    & 3800 \\
& Number of correctly selected samples & 2874      & 35251    & 3549 \\
& Accuracy of selected samples (\%)      & 79.88     & 97.85   & 93.39 \\
\hline
\multirow{3}{*}{Classifier and discriminator confidences together} & Number of selected samples       & 3557      & 35123    & 3558 \\
& Number of correctly selected samples & 2873      & 34558    & 3454 \\
& Accuracy of selected samples (\%)      & 80.77     & 98.39   & 97.08 \\
\bottomrule
\end{tabular}}
\end{table*}

\begin{table}[ht]
\renewcommand{\arraystretch}{1.3}
\caption{Ablation experiments.}
\label{table:ablation}
\vspace{2mm}
\centering
\scalebox{1.07}{
\begin{tabular}{@{}l c c c >{\columncolor[gray]{0.8}}c}
\toprule
    Method & \rotatebox{90}{Bicycle } & \rotatebox{90}{Car } & \rotatebox{90}{Person } & \rotatebox{90}{\textbf{Average }}\\
\hline
SGADA-Cls & 87.36 & 95.27 & 90.62 & 91.08\\
SGADA-Cls+Disc & 87.13 & 94.44 & 92.03 & \textbf{91.20}\\
\bottomrule
\end{tabular}}
\end{table}

Per-class domain adaptation performances are reported in Table \ref{table:results}. The results show that our proposed method which uses classifier and discriminator confidences together for self-training outperforms the state-of-the-art methods. Although DTA \cite{dta} and DANN \cite{dann} perform well for \textit{car} and \textit{person} classes respectively, their performance  for \textit{bicycle} class cannot reach the top performances since the number of samples for \textit{bicycle} class is much less than the other classes. On the other hand, our proposed method achieves more balanced performance scores for all classes and outperforms other methods. It is important to address this problem since datasets for real-world problems usually include imbalanced classes \cite{buda2018systematic, classimbalance}. As shown in the table, our proposed method achieves more balanced class-wise accuracies compared to our base method ADDA \cite{adda}, and furthermore our method increases the average accuracy over our base method.
\vspace{-2mm}

\paragraph{Qualitative Analysis.} We visualize the feature representations on target thermal domain with t-SNE \cite{tsne} for qualitative analysis in Figure \ref{fig:tsne}. The features of source only model on target domain can not be discriminated very well while ADDA \cite{adda} discriminate some overlapping points in the feature space. Our proposed model which uses only classifier confidences for self-training (SGADA-Cls) learns more discriminative representations. As shown in the figure, our proposed model which uses classifier and discriminator confidences together for self-training (SGADA-Cls+Disc) further enlarges inter-class distances, especially for \textit{car} and \textit{person} classes.
\vspace{-2mm}

\paragraph{Ablation Study.} To evaluate the contributions of our proposed method, we perform ablation studies. We examine effects of using classifier and/or discriminator confidences for pseudo-label selection. As described in Section \ref{sec:method}, we select samples using our base method. Table \ref{table:confidence} shows three cases where we select target domain samples using only the confidences of the classifier, using only the confidences of the discriminator, and using the confidences of both the classifier and the discriminator. As shown in the table, if we utilize the discriminator confidences, the number of selected samples for the \textit{person} class decreases. Moreover, when we use both discriminator and classifier confidences to select target domain samples, accuracy of the pseudo-labels increases significantly for all classes. This results in better separation of feature representations as depicted in Figure \ref{fig:tsne} (c)-(d). Furthermore, since accuracy of selected samples for all classes using classifier and discriminator confidences is higher than the other cases, class imbalance of our proposed method (SGADA-Cls+Disc) reduces compared to SGADA-Cls as shown in Table \ref{table:ablation}. And thus, the overall accuracy of our proposed method surpasses SGADA-Cls, resulting in the best overall performance.
\vspace{-2mm}

\section{Conclusion}

In this paper, we propose a self-training guided adversarial domain adaptation method in order to investigate the efficacy of combining visible spectrum and thermal image modalities by using unsupervised domain adaptation. To overcome the generalization problems of the current adversarial domain adaptation methods, we employ pseudo-labels obtained from a classifier trained on RGB images, and train our method with these pseudo-labels. In order to demonstrate results of our method, we use large-scale RGB dataset MS-COCO as the source domain and thermal dataset FLIR ADAS as the target domain. Quantitative and qualitative results show that our proposed method achieves better results than the state-of-the-art adversarial domain adaptation methods by learning more generalized feature representations for target thermal domain.

{\small
\bibliographystyle{ieee_fullname}
\bibliography{egbib}
}

\end{document}